\pdfoutput=1

\documentclass{article}
\usepackage{icassp}
\usepackage{amsmath,graphicx}

\usepackage{times}
\usepackage{latexsym}
\usepackage{todonotes}
\usepackage{graphicx}
\usepackage{multirow}
\usepackage[T1]{fontenc}
\usepackage{booktabs}

\usepackage[utf8]{inputenc}
\usepackage{inconsolata}

\usepackage{microtype}
\usepackage[font=footnotesize,labelfont=bf]{caption}
\usepackage{subcaption}

\usepackage{arydshln}
\title{Towards a World-English Language Model \\ for On-Device Virtual Assistants}
\name{\begin{tabular}{c}Rricha Jalota$^{\oplus, \star}$ $\qquad$ Lyan Verwimp$^\dagger$ $\qquad$ Markus Nussbaum-Thom$^\dagger$ \\ $\qquad$ Amr Mousa$^\dagger$ $\qquad$ Arturo Argueta$^\dagger$ $\qquad$ Youssef Oualil$^\dagger$\end{tabular}}
\address{$^\oplus$AppTek GmbH $\qquad$ $^\dagger$Apple}

\begin{document}
%
\maketitle

\def\thefootnote{*}\footnotetext{Work done while the author was an intern at Apple.}\def\thefootnote{\arabic{footnote}}

\begin{abstract}
Neural Network Language Models (NNLMs) for Virtual Assistants (VAs) are generally language-, region-, and in some cases, device-dependent, which increases the effort to scale and maintain them.
Combining NNLMs for one or more of the categories is one way to improve scalability. 
In this work, we 
combine regional variants of English 
to build a ``World English'' NNLM for on-device VAs. 
In particular, we investigate the application of adapter bottlenecks to model dialect-specific characteristics in our  existing production NNLMs {and enhance the multi-dialect baselines}.
We find that adapter modules
are more effective in modeling dialects than specializing entire sub-networks.
Based on this insight and leveraging the design of our production models, we introduce a new architecture for World English NNLM that meets the accuracy, latency and memory constraints of our single-dialect models. 
\end{abstract}


\section{Introduction}
In on-device Virtual Assistants, it is common to deploy distinct Automatic Speech Recognition (ASR) models optimized for each language, region, and device~\cite{nussbaum-thom-etal-2023-application}. 
This allows the models to better capture regional trends and dialects, while meeting hardware constraints. 
However, maintaining several models
requires a lot of effort, and shipping 
new features
requires testing 
all combinations of device and 
language variants.
Therefore, 
{building a joint model for serving all variants of a language can}
improve the scalability of this process by reducing the number of different recipes that need to be maintained, and 
also reduces the environmental cost of training several models.
In this work, we focus on the language model (LM) component of the hybrid ASR pipeline and build a World-English NNLM by combining three dialects of English that are spoken in the USA, UK and India, henceforth referred to as: en\_US, en\_GB, and en\_IN. Specifically, we focus on the LMs that can be stored and used for ASR inference on device. Hence, the use of models that do not fulfill this requirement (e.g., Large Language Models like GPT-3 \cite{gpt3}, etc.) is out of scope of this paper.

Earlier works studying multilingual ASR either focused on the Acoustic Model (AM) in hybrid ASR~\cite{yadav2022survey} or on an end-to-end (E2E) ASR architecture. Multilingual E2E ASR models 
either do not have external LMs~\cite{li2018multi, joshi2021multiple, pratap2020massively}, or if they have them,
the LMs are often trained on the pooled dataset without additional enhancements~\cite{yadav2022survey}. 

Recently, adapter modules have become a
popular architecture extension to improve Multilingual Acoustic Modeling~\cite{Winata2020AdaptandAdjustOT} and model language-specific traits in end-to-end ASR~\cite{kannan2019large}. 
Adapters are parameter-efficient modeling units consisting of a down-projection, followed by a non-linearity and an up-projection~\cite{rebuffi2017learning, houlsby2019parameter}. They are added either after every self-attention layer~\cite{Winata2020AdaptandAdjustOT} or feed-forward layer~\cite{houlsby2019parameter, pfeiffer2020mad} in the encoder/decoder block of Transformer-based architectures, and usually add around 5\% parameters to the model. To the best of our knowledge, 
Kannan et al.~\cite{kannan2019large} 
were the first to apply 
language-specific adapters in a non-attention based
RNN-T framework~\cite{graves2012sequence} for Multilingual Acoustic-Modeling.

Contrary to previous works, we investigate the application of adapters and compare different adapter training schemes 
in two distinct Feedforward LM architectures, based on the Fixed-size Ordinally-Forgetting Encoding (FOFE) method~\cite{zhang2015fixed,nussbaum-thom-etal-2023-application}. 
We prefer FOFE-based models over transformer-based models since they have better accuracy-latency trade-off 
for our two applications~\cite{nussbaum-thom-etal-2023-application}: 
Speech-to-Text (STT) and Assistant.
Speech-to-Text requests are dictated messages such as notes and e-mails, while Assistant covers VA requests from various domains such as music, timer, etc. 
In our use case, the dialect information is already set by the user and thus known beforehand. It is used to train dialect-specific modules and to enable the activation of the relevant sub-network during inference.



As opposed to previous works~\cite{Winata2020AdaptandAdjustOT,kannan2019large} that focus on multilingual modeling to improve accuracy on one or more low-resource languages, in this paper,
we aim to build a multi-dialect model, wherein each dialect can be considered high-resourced. 
Our contributions are the following:
(1) we show 
that the accuracy gains of adapters are also applicable to our FOFE-based architectures, 
(2) we conduct an in-depth analysis on the placement, 
training strategies and variants of adapters in FOFE-based NNLMs, and
   (3) we introduce a new adapter-based model that leverages the design of our FOFE-based architectures and meets the accuracy, latency and memory constraints of on-device VAs.
\section{Model Architecture}
First, we briefly describe the existing FOFE-based single-dialect architectures. Then, we define World-English baselines and present ways to enhance them by adding adapters, ending with the introduction of the new adapter-based architecture.

\subsection{Baseline FOFE-based NNLMs}
\label{sec:fofe}
The FOFE method~\cite{zhang2015fixed} 
uniquely encodes word-order information using a recursive formula with a forgetting factor, thereby enabling a feedforward neural network (FNN) to model long-term dependencies.
To jointly handle both Assistant and STT applications, \cite{nussbaum-thom-etal-2023-application} extend FOFE-FNNs to two architectures,
namely, \textbf{Mixture FOFE} and \textbf{Application-Dependent (AD) FOFE}.
Mixture FOFE, shown in Figure~(\ref{fig:mixture}) consists of a word embedding layer (which for all architectures is shared with the output), followed by a FOFE layer and 
$N$ parallel blocks of stacked feedforward layers,
one of which acts as
an unsupervised mixture~\cite{nussbaum-thom-etal-2023-application}. The unsupervised mixture provides weights to average out the 
features learned by 
the other $N-1$ parallel blocks
and then the averaged output is fed to the projection layer.

Unlike Mixture FOFE, 
in AD FOFE,
distinct sub-networks are trained for each application - in our case Assistant and STT.
Each sub-network consists of a stack of $L$ feedforward layers and carries its own
projection head and output bias. 
During training, the application-based sub-networks are jointly trained, while during inference, only the sub-network pertaining to the application remains active. This gives a smaller memory footprint to the AD FOFE model and allows faster inference. 
For the World-English model, we consider both Mixture FOFE and AD FOFE as our base architectures and investigate the most optimal setup.
\subsection{World-English NNLMs}
\label{sec:worlden}
%
\begin{figure*}[!htp]
\centering
\begin{subfigure}{0.24\textwidth}
 \centering
 \includegraphics[width=\textwidth]{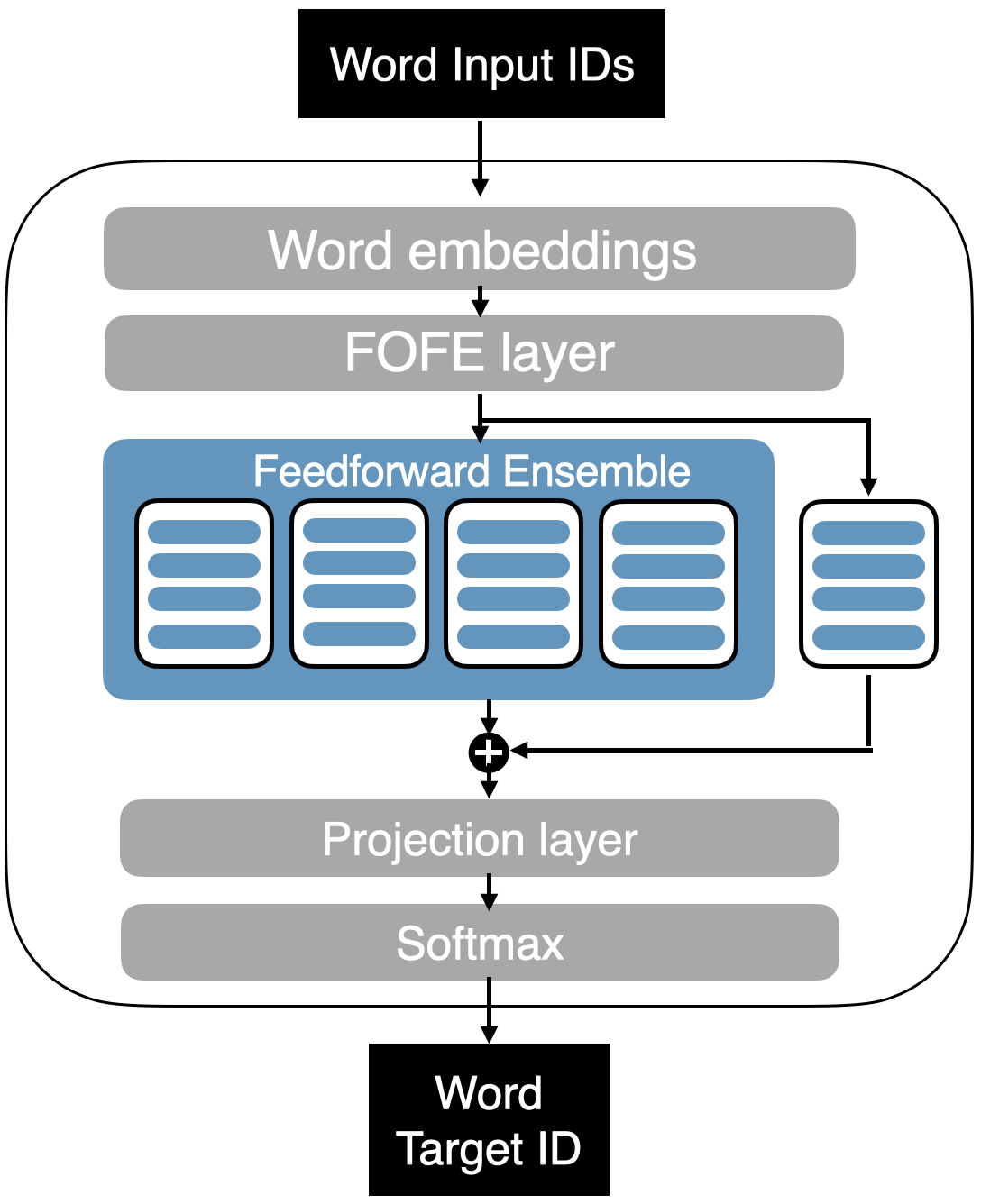}
 \caption{Mixture}
 \label{fig:mixture}
\end{subfigure}
\begin{subfigure}{0.24\textwidth}
 \centering
 \includegraphics[width=\textwidth]{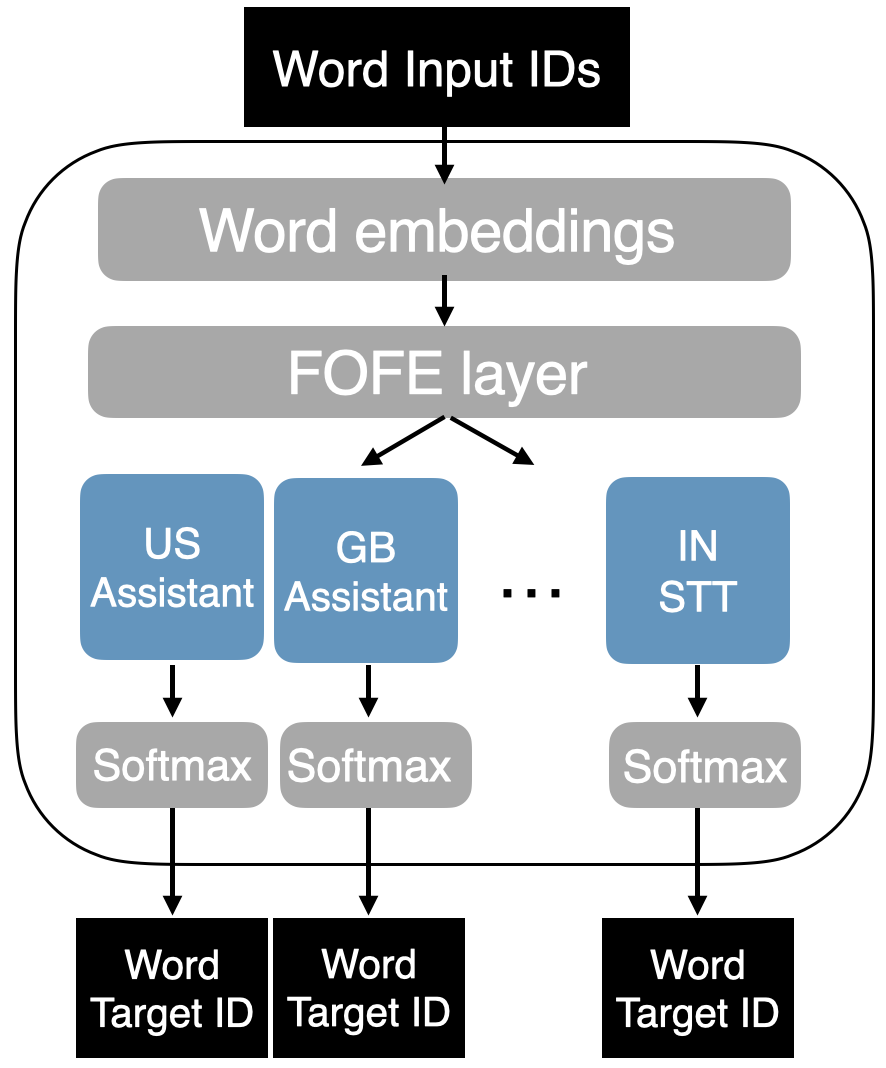}
 \caption{AD}
 \label{fig:clsdep}
\end{subfigure}
\begin{subfigure}{0.24\textwidth}
 \centering
 \includegraphics[width=\textwidth]{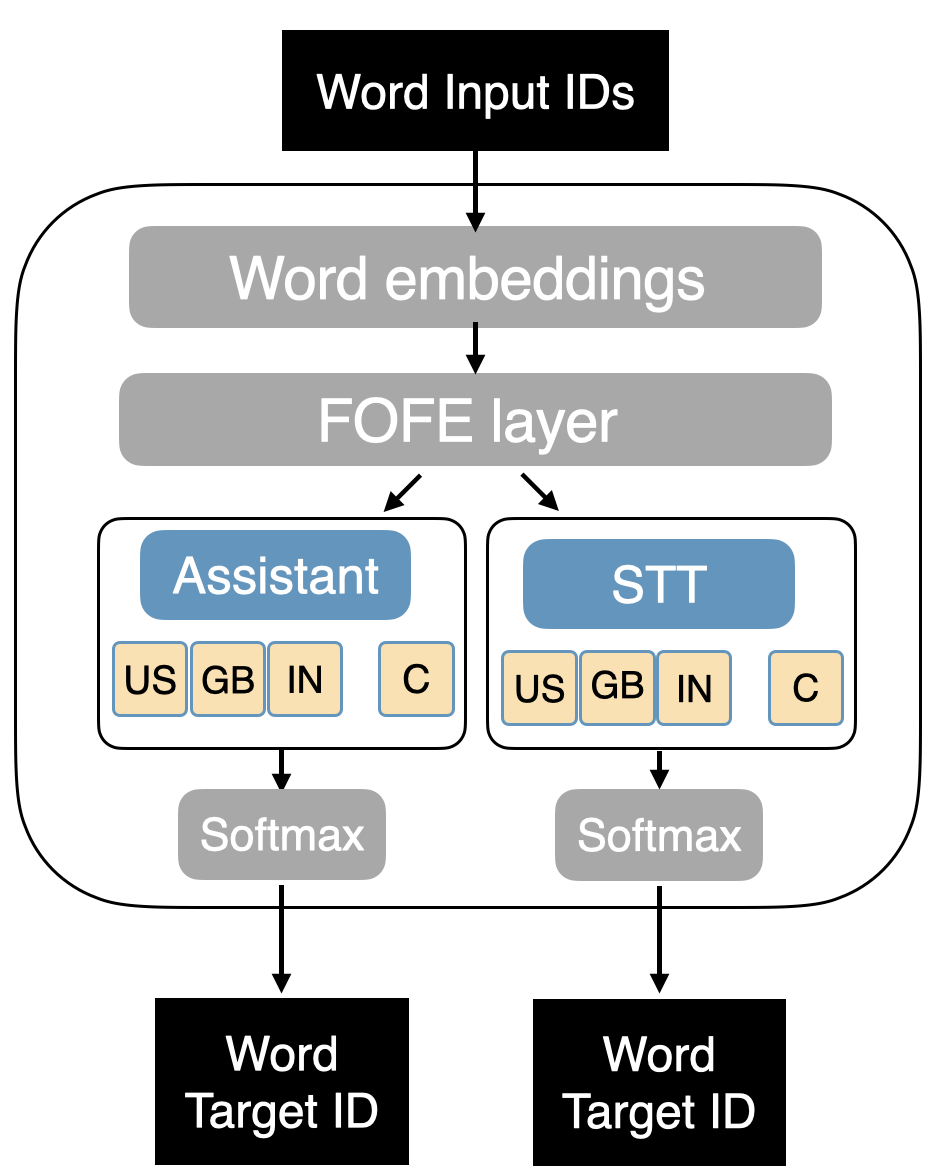}
 \caption{AD + DA}
 \label{fig:dualcls}
\end{subfigure}
\begin{subfigure}{0.24\textwidth}
 \centering
 \includegraphics[width=\textwidth]{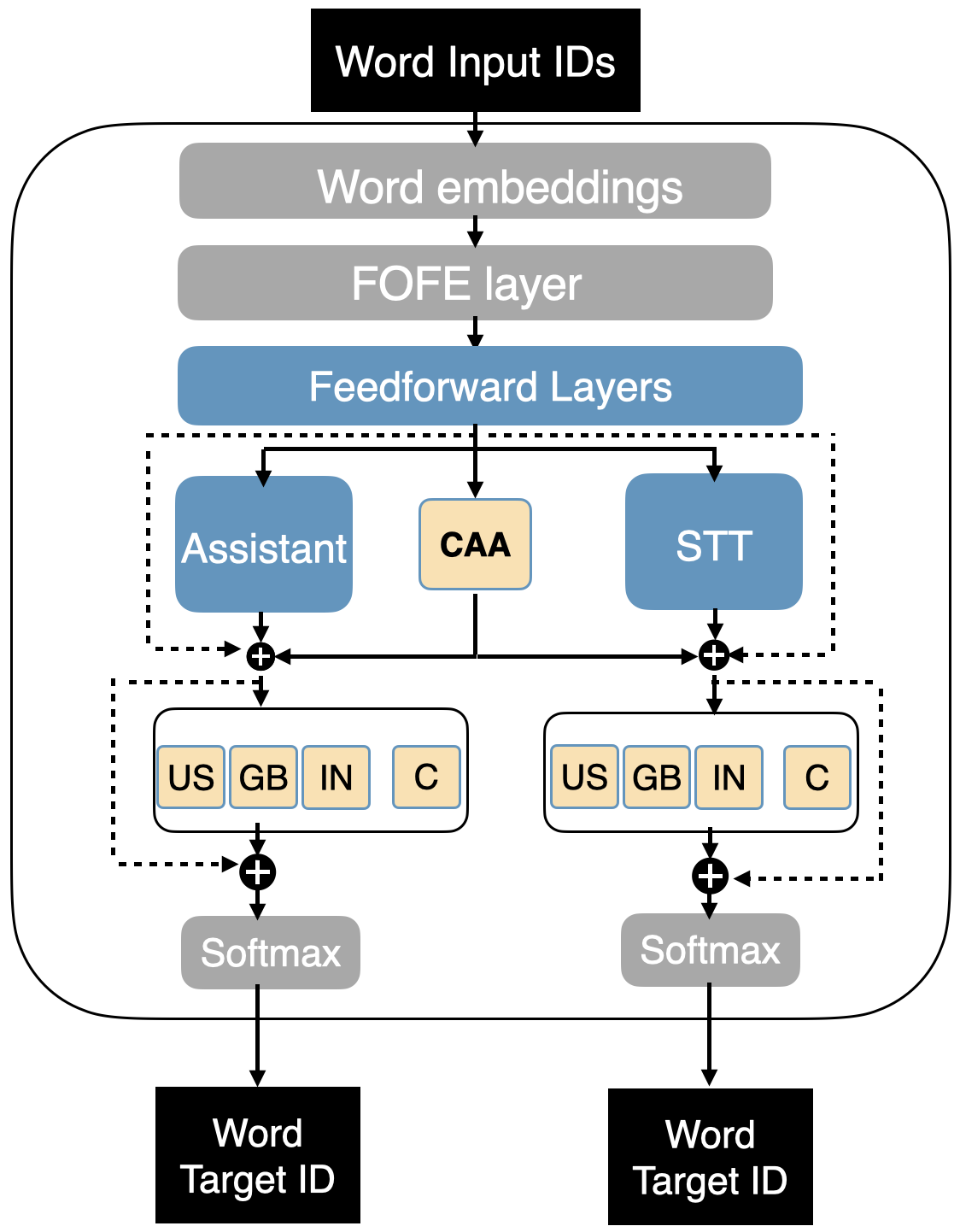}
 \caption{AD + CAA + DA }
 \label{fig:newarc}
\end{subfigure}
\caption{FOFE-based NNLM Architectures. The components in blue denote feedforward layers. US, GB, IN refer to American, British and Indian English. The abbreviation C in figures~\ref{fig:dualcls} and \ref{fig:newarc} refers to the Common Dialect Adapter and CAA refers to Common Application Adapter. Figure~(\ref{fig:mixture}): Mixture FOFE model, (\ref{fig:clsdep}): Multi-dialect AD FOFE (\texttt{AD}), (\ref{fig:dualcls}): AD FOFE with Dual Adapters (\texttt{AD+DA}) and (\ref{fig:newarc}): AD FOFE with CAA and Dual Adapters (\texttt{AD+CAA+DA}).}
\label{fig:models}
 \vspace*{-0.5cm}
\end{figure*}
We first establish the baselines for World-English NNLM by feeding the Mixture FOFE and AD FOFE models with multi-dialect data (curation explained in Section~\ref{sec:setup}). While the baseline Mixture FOFE model does not change with the number of dialects, in case of AD FOFE, the number of sub-networks increases by two with the addition of each dialect, as shown in Figure~(\ref{fig:clsdep}). This means, the baseline World-English AD FOFE model consists of six sub-networks, with each sub-network modeling an application in the given dialect. 

\medskip
\noindent{{\textbf{Extension with Adapters}}}: Following previous works~\cite{doddapaneni2021primer,kannan2019large} that use adapters as an alternative to fine-tuning, 
we inspect if adapters can bring similar accuracy gains to our pre-trained World English baseline models. Similar to~\cite{houlsby2019parameter}, we define an adapter as a bottleneck, where the original dimension $d$ is projected onto a smaller dimension $k$, followed by a non-linearity (ReLU) and projected back to $d$ dimensions. The adapter output is then combined with the residual from the layer preceding the adapter module.

\medskip
\noindent{\underline{\textit{Adapter placement:}}} The architectural addition of adapters is not trivial~\cite{pfeiffer-etal-2021-adapterfusion}, therefore, we first investigate their placement. In Mixture FOFE (see Fig.~(\ref{fig:mixture})), a dialect-specific adapter can be placed at various positions: (i) before the projection layer, (ii) on top of the last hidden layer in each of the parallel blocks,
(iii) a combination of (i) and (ii), (iv) after each hidden layer in each of the parallel blocks, and (v): combination of (iv) and (i). 

\medskip
\noindent{\underline{\textit{Training Strategies:}}} Next, we examine three different adapter training schemes. Adapters are usually trained in a two-step process~\cite{houlsby2019parameter, pfeiffer2020mad, kannan2019large, Winata2020AdaptandAdjustOT}. First, a base model is pre-trained on the combined data from all dialects. In the second step, all the model parameters are frozen, and the adapter is added, 
which is trained on the dialect-specific data. Since the adapter is randomly initialized to mimic a near-identity function before learning the dialect-specific weights, we call this training setup \textbf{Randomly-Initialized Adapter} \textbf{(\texttt{RI-A})}. In the second training strategy, we include the adapter in the architecture in the first step (similar to the baseline in \cite{DBLP:conf/naacl/PfeifferGLLC0A22}) and train the entire network with the multi-dialect data. We call this \textbf{Adapter-Pretraining} \textbf{(\texttt{PT-A})}. In the third training scheme, we fine-tune the pretrained adapter with dialect-specific data and therefore, refer to it as \textbf{Adapter-Finetuning} \textbf{(\texttt{FT-A})}. {We hypothesize that adapters starting from pre-trained weights would converge faster and might perform better than \texttt{RI-A}.}

\medskip
\noindent{\underline{\textit{Adapter Variant:}}} Finally, we inspect 
an adapter variant, namely, \textbf{Dual adapters} \textbf{(\texttt{DA})}~\cite{Winata2020AdaptandAdjustOT} in AD FOFE, which besides a dialect-specific adapter also contains a common dialect adapter (C) to learn a shared representation of dialects.
We integrate dual adapters in AD FOFE by reducing the number of sub-networks to the number of applications.
As shown in Figure~(\ref{fig:dualcls}), instead of having a dedicated sub-network for each application per dialect, we have only two sub-networks (one for each application) and add dual adapters on top of them, thereby reducing the number of parameters.  

\medskip
\noindent{{\textbf{Proposed Architecture}}}:
It is observed in \cite{nussbaum-thom-etal-2023-application} that for monolingual setups, although AD FOFE is preferred for its lower latency, Mixture FOFE is more accurate\footnote{In this work, accuracy refers to word error rate (WER) while performance refers to latency.}. We hypothesize, this higher accuracy might be due to the shared representation of applications in Mixture FOFE. Hence, to enhance the accuracy of AD FOFE while preserving its low latency, we introduce a novel architecture that combines the AD FOFE and dual adapters with Mixture FOFE (see Figure~(\ref{fig:newarc})). \\
Following the FOFE layer, we add a block of $L$ feedforward layers to enable joint learning of applications in all three dialects. This is followed by application-dependent sub-networks and a Common Application Adapter (CAA), which is added in parallel to the sub-networks. Similar to the common dialect adapter in dual-adapters, CAA would further facilitate learning application-agnostic traits. The combined output from the application-sub-networks, CAA and the residual from the block of feedforward layers is directed to
dual-adapters, placed atop each sub-network for modeling dialects. The architecture concludes with output from application-specific projection heads and is henceforth referred as, Application-Dependent model with Common Application Adapter and Dual Adapters (\textbf{\texttt{AD+CAA+DA}}).
%
\vspace*{-0.3cm}
\section{Experimental Setup}
\label{sec:setup}
\noindent{\textbf{Data:}} Our training data comes from anonymized and randomly sampled user requests from several domains (media, photos, calendar, etc.) and applications (Assistant and STT) in each of the three dialects: en\_US, en\_GB, and en\_IN. For each dialect, the relevance weights for the data sources are estimated using the same technique as described in \cite{nussbaum-thom-etal-2023-application}.
Given that all the three dialects are high-resourced, sampling equal amounts of data for each dialect turned out to be the most optimal choice for our use case. \\

\noindent{\textbf{Training:}} 
We train single-dialect NNLMs on 12B words with the top 100k most frequent words as vocabulary. For the multi-dialect setup, we sample the training data as explained above and set the training words to 36B and vocab size to 150k. The vocab size is empirically chosen on the development set such that the coverage with respect to the single-dialect vocabularies is greater than 75\% while still achieving meaningful WER reductions and shorter training times.
For inference, the multi-dialect NNLM is fed into distinct, dialect-specific ASR systems. This means, except the NNLM, all other components of the system are specific to the dialect and remain intact. Similar to \cite{nussbaum-thom-etal-2023-application}, we set the values of $N$ and $L$ in our models to $5$ and $4$, respectively. 
The rest of the hyperparameters for training the FOFE-based models are the same as reported in \cite{nussbaum-thom-etal-2023-application}. \\
\noindent{\textbf{ASR System:}} The ASR System consists of a CNN-acoustic model~\cite{huang2020sndcnn} and a FOFE-based NNLM~\cite{nussbaum-thom-etal-2023-application} that is used in the first pass.  \\

\noindent{\textbf{Evaluation:}} The models are evaluated in terms of accuracy, size (measured by the number of parameters to account for memory constraints), and on-device latency. We estimate latency using \textit{ASR Processing Latency}, which is defined as the time from when the user stops speaking to when the ASR posts the final result. Both average and 95th Percentile (P95) results are reported, based on an average of 3 runs on device.   
We evaluate accuracy using Word Error Rate (WER) on three test sets: 
Assistant (Ast.), 
Speech-to-Text (STT) 
and 
Tail Entities (T.E.). Assistant and STT consist of general VA requests sampled from the actual distribution, thereby containing the most frequent (head-heavy) queries. Tail Entities is a test set synthesized with Text-to-Speech, containing less-frequently occurring queries with tail entities.  Table~\ref{tab:stats} presents the development and test set statistics.
\vspace*{-0.1cm}
\begin{table}[!htb]
\centering
\small
\scalebox{0.8}{
\begin{tabular}{ccccc} \hline
\textbf{Split} &  \textbf{Dialect}       & \textbf{Ast.} & \textbf{STT} & \textbf{T.E.} \\ \hline
\multirow{3}{*}{dev}  & {en\_US} & {215,299} & {285,853} & {-}  \\  
      & {en\_GB}  & {148,814} & {111,650} & {-}       \\ 
      & {en\_IN}  & {145,795}  & {55,907}   & {-}       \\ \hline
\multirow{3}{*}{test}  & {en\_US}   & 226,371          &    292,477          &  454,159  \\
   & {en\_GB}  & 155,232            &    114,103        &          232,285         \\
    &  {en\_IN}  & 153,862       &   54,562       &  239,852 \\ \hline          
\end{tabular}}
\caption{{Number of words development and test sets.}}
\label{tab:stats}
\vspace*{-0.3cm}
\end{table}
\section{Results}
\label{sec:results}
\noindent{\textbf{Adapter Placement:}} 
We perform Bayesian Optimization on the development set to find the most optimal placement and compression dimension $k$ (128, 96, 48) starting from a hidden dimension, $d$ = 768.
We observe that adding only a single dialect-specific adapter before the projection layer, i.e. placement (i) in Sec.~\ref{sec:worlden}, with a compression dimension of 96 ($<$ 0.5\% more parameters)
is more effective than adding multiple adapters to the architecture. This is in contrast to the  previous works~\cite{houlsby2019parameter, pfeiffer2020mad, Winata2020AdaptandAdjustOT, DBLP:conf/naacl/PfeifferGLLC0A22}, where adapters are added in every block of the encoder/decoder.

\medskip
\noindent{\textbf{Training Strategies:}} 
We then compare the three adapter training strategies in Mixture FOFE to verify if one could be preferred over another. We observe that the results are mixed and vary with test sets across dialects. One would expect that fine-tuning the pre-trained adapter (\texttt{FT-A}) shows further improvements over \texttt{PT-A} and \texttt{RI-A} across dialects. However, this does not hold true across all test sets.
Since the improvements from \texttt{PT-A} are consistent across all dialects on an average, we adopt this training strategy in all experiments with adapters reported in Table~\ref{tab:word-eng-part}.

\medskip
\noindent{\textbf{Adapters in Multilingual FOFE models:}} 
In Table~\ref{tab:word-eng-part}, we report the WERs of the best performing multi-dialect FOFE models combined with adapters using the optimal placement and training strategy. Firstly, we observe that both the models, Mixture (\texttt{Mix}) and \texttt{AD}, already have good accuracy compared to the single-dialect baselines (\texttt{Mono}), with \texttt{Mix} outperforming \texttt{AD} in most cases.
Adding adapters (\texttt{Mix+A}) gives slight but consistent improvements to \texttt{Mix}. 
However, for \texttt{AD+A}, 
the results are more mixed, e.g. the accuracy on tail entities improves by 11.6\% for en\_IN while for en\_GB it degrades w.r.t. \texttt{AD}.
However, given that AD models are smaller and faster in inference~\cite{nussbaum-thom-etal-2023-application}, we want to further bridge the accuracy gap w.r.t. Mixture models.

\begin{table}[!htb]
\resizebox{1\columnwidth}{!}{
\centering
\begin{tabular}{@{}c||c|ccc|ccc|ccc@{}}
\hline
 \multirow{2}{*}{\textbf{Model}} & \multirow{2}{*}{\shortstack[c]{\textbf{Model} \\ \textbf{Size}}}
   & \multicolumn{3}{|c|}{\textbf{en\_US}}           & \multicolumn{3}{c|}{\textbf{en\_GB}}          & \multicolumn{3}{c}{\textbf{en\_IN}}  \\ 
  \cline{3-11}
    &          & \textbf{Ast.}  & \textbf{STT} & \textbf{T.E.}       & \textbf{Ast.}  & \textbf{STT} & \textbf{T.E.}       & \textbf{Ast.}     & \textbf{STT}      & \textbf{T.E.}       \\
    \hline
\texttt{Mono} &  111M  & 3.97          & 3.47          & 18.24 & \textbf{5.26} & \textbf{6.16} & 16.3 & 6.92          & 9.62 & 26.14 \\ \hline
\texttt{Mix}  &   89M         & 3.97          & \underline{3.41}          & 16.84 & \underline{5.33}          & \underline{6.17} & 16.29         & \textbf{6.69} & 9.46 & 24.01 \\ 
\texttt{Mix+A} &  89M & \underline{3.95}  &  \underline{3.41}     & \textbf{16.83} & \underline{5.33}       & 6.18     & \textbf{16.27} & \textbf{6.69} & \textbf{9.18} & \underline{23.99} \\ \hline
\texttt{AD}   &  54M     &   4.01          & 3.43          & 17.52 & \underline{5.34}          & 6.28 & 16.69         & 7.16          & 9.57 & 24.67 \\ 
\texttt{AD+A}  &  55M  & 3.99          & {3.41} &      21.94         & 5.38          & 6.33 & 21.88             &     7.24  & 9.64       & \textbf{21.80}       \\ 
\texttt{AD+DA} &  45M  & 3.97     & 3.42          & \underline{17.32} & 5.36          & \underline{6.21} & 16.53         & \underline{6.90}          & 9.54 & 24.34 \\ 
\texttt{AD+CAA+DA}  &  49M   & \textbf{3.93} & \textbf{3.39} & \underline{17.32} & 5.35          & 6.25 & \underline{16.44}         & \underline{6.90}          & \underline{9.42} & 24.32 \\ 
\hline
\end{tabular}}
\centering
\caption{First-pass decoding results (WERs) of (i) the best (Mixture FOFE/AD FOFE) single-dialect model \texttt{Mono}; (ii) Multi-dialect Mixture FOFE (\texttt{Mix}) with (iii) Pre-trained Adapter (\texttt{Mix+A}); (iv) Multi-dialect AD FOFE (\texttt{AD}) with (v) Adapter (\texttt{AD+A}), (vi) Dual-Adapter (\texttt{AD+DA}), and (vii) with Common Application Adapter \texttt{(AD+CAA+DA)}. The second column shows the aggregated model size across dialects. \textit{The best WERs within the model families are underlined, while bold numbers highlight the best overall result.}}
\label{tab:word-eng-part}
\end{table}
\vspace*{-0.2cm}
\noindent{\textbf{Adapter Variants:}} 
In \texttt{AD+A}, only the word embeddings benefit from parameter sharing across dialects and applications. This might be the reason behind the mixed results. We try to overcome the degradation on tail entities by introducing more shared parameters. In \texttt{AD+DA}, the number of sub-networks is reduced to two, and dual-adapters are added to characterize the dialect-specific and dialect-agnostic traits. This not only reduces the model size by almost $10\%$ but also relatively improves the baseline \texttt{AD} model by an average of $1.18\%$
on test sets across dialects (see lower half of Table~\ref{tab:word-eng-part}).

\medskip
\noindent{\textbf{Proposed Architecture:}}
Finally, we investigate if the proposed architecture improves accuracy over AD FOFE variants. 
As expected, adding a shared representation for applications in AD FOFE (i.e., \texttt{AD+CAA+DA}) relatively improves it on all test sets
by an average of $1.41\%$ over \texttt{AD} and 
marginally over \texttt{AD+DA}.
In fact, for en\_US, \texttt{AD+CAA+DA} also marginally outperforms the Mixture FOFE variants on head queries. However, 
\texttt{Mix+A} still 
achieves a better WER 
across most test sets due to its larger model size and thus, larger shared representation among dialects and applications.
Overall, both the multi-dialect models: \texttt{Mix+A} and \texttt{AD+CAA+DA}, improve the single-dialect baselines on all dialects by an average of $1.41\%$ and $1.63\%$ on head-heavy test sets, and $5.38\%$ and $3.72\%$ on tail entities, respectively.
In terms of accuracy, \texttt{Mix+A} is the best choice for World-English NNLM. However, it is $45\%$ bigger in size than \texttt{AD+CAA+DA}.
 
\begin{table}[!htb]
\centering
\small
\scalebox{0.8}{
\begin{tabular}{ccccc} \hline
\textbf{Model}       & \textbf{Ast. Avg.} & \textbf{Ast. P95} & \textbf{STT Avg.} & \textbf{STT P95} \\ \hline
    \texttt{Mono\_150k}  & 334 & 425 & 50 & 185 \\ 
    \texttt{Mix+A}   & 421  & 785 & 74 & 230  \\
    \texttt{AD+CAA+DA}  & 359 & 474 & 54  &  182 \\ \hline          
\end{tabular}} 
\caption{{Latency Results (in milliseconds) on an average of 3 runs on device. \texttt{Mono\_150k} refers to single-dialect AD FOFE with 150k vocab.}}
\label{tab:latency}
\vspace*{-0.2cm}
\end{table}
Next, we compare the models \texttt{Mix+A} and \texttt{AD+CAA+DA} in terms of latency on en\_US test sets. As increasing the vocab size results in increased latency, to make the comparison more fair, we compare our multidialect models to single-dialect AD FOFE with 150k vocab size (\texttt{Mono\_150k}).
Due to random fluctuations on device, relative changes in latency less than $10\%$ are considered to be equally fast.
Results in Table~\ref{tab:latency} show that for both applications, \texttt{AD+CAA+DA} matches the latency of \texttt{Mono\_150k}.
Furthermore, it outperforms \texttt{Mix+A} by an average of $27\%$ on STT and is even $40\%$ faster on the top-5\% queries from Assistant that incur the largest latency (P95). 

In summary, our proposed architecture \texttt{(AD+CAA+DA)} for World-English NNLM offers a favorable accuracy-latency-memory trade-off, showcasing its potential for deployment.
\section{Conclusion}
We build a World-English NNLM for an on-device ASR system, starting with three high-resourced English dialects. We first examine the application of adapters in FOFE-based architectures. Based on our findings, we introduce a new architecture 
to bridge the accuracy gap between the baseline Mixture FOFE and AD FOFE models. This model relatively improves the accuracy of single-dialect baselines by an average of $1.63\%$ on head-heavy test sets and $3.72\%$ on tail entities across dialects. Moreover, it matches the
latency and memory constraints of on-device VAs, which indicates that all single-dialect baselines can be replaced by this single model. In the future, the insights from our experimental results will be leveraged to truly achieve a World-English NNLM spanning all dialects.
\vspace{-0.4cm}
\clearpage
\bibliographystyle{IEEEbib}
\bibliography{anthology,custom}

\end{document}